# Solution Methods for Constrained Markov Decision Process with Continuous Probability Modulation

**Janusz Marecki,   Marek Petrik,   Dharmashankar Subramanian**
Business Analytics and Mathematical Sciences
IBM T.J. Watson Research Center
Yorktown, NY
{marecki,mpetrik,dharmash}@us.ibm.com

## Abstract

We propose solution methods for previously-unsolved constrained MDPs in which actions can continuously modify the transition probabilities within some acceptable sets. While many methods have been proposed to solve regular MDPs with large state sets, there are few practical approaches for solving constrained MDPs with large action sets. In particular, we show that the continuous action sets can be replaced by their extreme points when the rewards are linear in the modulation. We also develop a tractable optimization formulation for concave reward functions and, surprisingly, also extend it to non-concave reward functions by using their concave envelopes. We evaluate the effectiveness of the approach on the problem of managing delinquencies in a portfolio of loans.

## 1   Introduction

This paper is motivated by the need of a loan services provider to efficiently manage a portfolio of loans in various, finite, *levels of delinquency* over a finite number of *decision periods*. In the absence of interventions, a loan is assumed to transition from one delinquency level to another across time periods according to an exogenous *base* transition probability. This transition probability can, however, be controlled by taking various intervention actions, the cost of which depends on the deviation from the base transition probability. The overall objective in managing such a portfolio of loans is to choose interventions that maximize the expected financial gain of a loan servicing operator (or equivalently to minimize its loan servicing cost), subject to some constraints on the performance of the loan portfolio in expectation. These performance constraints are motivated by both regulatory and business reasons, and are typically in terms of acceptable bounds on the *expected* percentage of loans that would result in a default (the most delinquent level) at the end of a planning horizon, or at various intermediate time periods. While we focus specifically on loans, our models and results are applicable to other domains, such as maintenance scheduling, debt collection, and marketing [1].

To determine the right sequence of such interventions, one needs to solve a stochastic dynamic decision problem. Note that it suffices to optimize the sequence of interventions independently for each loan, since all important metrics (decision-making objectives and constraints) are expressed in terms of expectations. For each decision period $t$ we assume a finite set of states $\mathcal{S}_t$ that represent the various levels of loan delinquency for the period $t$. For any loan state $s_t \in \mathcal{S}_t$, let $b(s_t)$ denote the base transition probability distribution over the finite support $\mathcal{S}_{t+1}$. The decision-maker can modify $b(s_t)$ into any probability distribution $p(s_t)$ that belongs to a set $\mathcal{P}_{s_t}$ of feasible distributions. In other words, $p(s_t)$ is the modulated transition probability to other delinquency states $\mathcal{S}_{t+1}$ after an intervention corresponding to $s_t$. The cost of achieving this modulation is assumed to be a function of the difference, $p(s_t) - b(s_t)$.

Given the chosen interventions, let $d(s)$ represent the probability of visiting state $s \in \mathcal{S}_T$ in time period $T$ following a sequence of $T-1$ interventions. In vector notation, we have that:

$$d = \alpha^\mathsf{T} P_1 \cdot P_2 \cdot \dots \cdot P_{T-1},$$

where $\alpha$ is an initial probability distribution over the finite set $\mathcal{S}_1$ and $P_t = [p(s_t)]_{s_t \in \mathcal{S}_t}$ is the transition probability matrix induced by the interventions. The portfolio performance constraints require that for some selected states $s$ and values $q(s)$, $d(s) \le q(s)$. Note that $d$ is a complex polynomial function of the decisions $p$. Consequently, the total expected costs corresponding to a sequence of $T-1$ interventions and transitions, as well as the performance constraints are also non-convex polynomials of degree $|T-1|$. Because non-convex polynomial optimization problems are usually very hard to solve, this direct formulation is unlikely to lead to a tractable solution.

To derive tractable algorithms we instead cast the problem as an instance of a *constrained* Markov decision process (CMDP) [2]. The MDP states in this formulation represent the levels of a loan delinquency and the actions represent the available interventions. The performance constraints can then be conveniently represented in the CMDP framework. While CMDPs with small state and action sets can easily be formulated and solved as linear programs, the loan delinquency management problem has a continuous action set—the available interventions can continuously adjust the transition probabilities between different states. Continuous action MDPs and CMDPs have been studied extensively in terms of existence of the optimal policies, but there have been few practical computational methods proposed [2, 9]. In this paper, we propose and analyze methods for solving some specific classes of CMDPs with continuous action sets.

Continuous action spaces in the form of compact spaces $\mathcal{P}_s$ have been considered in the context of robust Markov decision problems [4, 5]. Our setting is more complex because of the constraints on state visitation probabilities and non-linear reward functions. Continuous action spaces have also been considered in the context of reinforcement learning [6, 11]. The reinforcement learning approaches, unlike the methods we propose, are only approximate and cannot easily handle state probability constraints. Finally, continuous action spaces have been also considered in recent work on Markov decision process with linear transition structure [8, 7]. However, the required linear structure is not present in the loan servicing problem. Finally, CMDPs have recently been used in optimizing the tax collection for NY state [1]. The number of actions available in the tax collection problem, however, is small and the problem can be solved using standard MDP and reinforcement learning techniques.

The remainder of the paper is organized as follows. In Section 2, we define the *finite-horizon* MDP framework with continuous action spaces and state probability constraints. In Section 3, we show that the continuous-action CMDP can be reformulated as an identical finite action CMDP under some mild assumptions. While this formulation has a finite number of actions, it may still be too large to be solved efficiently in practice. In Section 4, we show a tractable formulation of the CMDPs with *concave* reward functions as a convex mathematical optimization program. Then, Section 5 extends the convex formulation to non-concave reward functions with tractable concave envelopes. Finally, Section 6 demonstrates the efficiency of the method on a realistic loan servicing problem.

## 2  Framework

In this section, we first describe the basic properties of constrained Markov decision processes with continuous modulation of transition probabilities. Then, we briefly discuss a CMDP formulation of the loan management problem.

We use $\Delta^n$ to denote the probability simplex in $\mathbb{R}^n$: $\Delta^n = \{p \in \mathbb{R}^n : \mathbf{1}^\mathsf{T} p = 1\}$—this represents the set of all probability distributions over $n$ elements. We also use $\mathbf{0}, \mathbf{1}, \mathbf{I}$ to denote a vector of all zeros, all ones, and an identity matrix respectively; their sizes are given by the context.

First, we define an abstract *finite-horizon constrained Markov decision process* (CMDP) $\mathcal{M}$ with continuously modulated transition probabilities. The finite time horizon is assumed to be: $t = 1 \ldots T$.

The finite state set at time $t$ is denoted as $\mathcal{S}_t$ and the set of all states is $\mathcal{S} = \bigcup_{t=1 \ldots T} \mathcal{S}_t$. The underlying base transitions probability from any state $s_t \in \mathcal{S}_t$ is $b(s_t) \in \Delta^{|\mathcal{S}_{t+1}|}$; that is the vector of transition probabilities from some $s_t$ to any $s_{t+1} \in \mathcal{S}_{t+1}$, when no action is taken. The infinite continuous actions space for any $s_t \in \mathcal{S}_t$ is denoted as $\mathcal{A}(s_t)$. The set $\mathcal{A}(s_t)$ must be *compact* and satisfies $\mathcal{A}(s_t) \subseteq \Delta^{|\mathcal{S}_{t+1}|}$ and $b(s_t) \in \mathcal{A}(s_t)$. The compactness assumption ensures that all the optima are achieved. An action $a_t \in \mathcal{A}(s_t)$ for $s_t \in \mathcal{S}_t$ denotes the modulated transition probability distribution over $s_{t+1} \in \mathcal{S}_{t+1}$.

The rewards are denoted as: $r(s_t, a)$ for state $s_t$ and action $a$. The initial probability distribution is: $\alpha \in \Delta^{|\mathcal{S}_1|}$. Finally, the solution must satisfy quality constraints such that the visitation probability for states in $\mathcal{Q}_i \subset \mathcal{S}$ are bounded by $q_i$ for some indices $i \in \mathcal{I}$.

Next, we summarize the known properties of the optimal solutions of CMDPs with continuous actions. Similarly to unconstrained MDPs, there exists an optimal Markov policy $\pi$ (e.g. Theorem 6.2 in [2]) under some mild assumptions, but this policy may need to be randomized. The set of randomized Markov policies $\Pi_R = \{\pi : \mathcal{S} \to \Delta^{|\mathcal{A}|}\}$. Note that the existence of an optimal policy requires that the action space is *compact*. A Markov policy is *deterministic* when the action distribution is degenerate; the set of deterministic policies is $\Pi_D = \{\pi : \mathcal{S} \to \mathcal{A}\}$.

**Definition 2.1.** The objective of the constrained MDP optimization is:

$$\max_{\pi \in \Pi_R} \mathbf{E}\left[\sum_{t=1}^{T-1} r(S_t, \pi(S_t))\right] \text{ s.t.} \sum_{\substack{t=1..T \\ s \in \mathcal{Q}_i}} \mathbf{P}\left[S_t = s\right] \leq q_i \, ,$$

for all $i \in \mathcal{I}$ where $S_t$ are state-$(\mathcal{S}_t)$-valued random variables and the constraints ensure the required solution quality.

*Remark* 2.2 (Uniformly optimal policies [2]). Unlike in regular MDPs, there may not be any uniformly optimal policies in a CMDP regardless of the initial state. The initial distribution is thus a key part of the CMDP definition.

In the remainder of the paper, we use sums instead of integrals to simplify the notation when using the continuous action space. Formally, one could replace all the sums by Lebesgue integrals.

For each policy $\pi \in \Pi_R$, $u_\pi(s_t, a) \in [0, 1]$ denotes joint *state action visitation probability*, and $d_\pi(s_t) \in [0, 1]$ denotes the *state visitation probability*. Using these terms, the return of a policy $\pi$ can be written as [2]:

$$\rho(\pi) = \sum_{t=1}^{T} \sum_{\substack{s_t \in \mathcal{S}_t \\ a_t \in \mathcal{A}(s_t)}} r(s_t, a_t) \cdot u_\pi(s_t, a_t) \quad (2.1)$$

where $u_\pi$ is uniquely determined by the following constraints [9]:

$$\sum_{a_t \in \mathcal{A}(s_t)} u_\pi(s_t, a_t) = d_\pi(s_t) \quad (2.2)$$

$$\sum_{s_t, a_t} u_\pi(s_t, a_t) \cdot a_t(s_{t+1}) = d_\pi(s_{t+1}) \quad (2.3)$$

$$d_\pi(s_1) = \alpha(s_1) \quad (2.4)$$

$$\frac{u_\pi(s_t, a_t)}{d_\pi(s_t)} = \pi(s_t, a_t), \quad (2.5)$$

where we implicitly assume that $s_t \in \mathcal{S}_t, a_t \in \mathcal{A}(s_t), a_{t+1} \in \mathcal{A}(s_{t+1})$ in (2.3) and the constraint must hold for each $t$ and $s_{t+1}$. Note that these constraint imply that $u \geq \mathbf{0}$.

The intuitive meaning of the above constraints is as follows. Constraint (2.2) requires that the state visitation probability is simply marginalized state-action visitation probability. Constraint (2.3) can be seen as a flow conservation constraint denoting that the probability of transiting to state $s_{t+1}$ from any state $s_t$ is equal to the probability of visiting the state. Note that $a_t$ in (2.3) is a vector of transition probabilities. Constraint (2.4) ensures that the initial probabilities are correct and finally, Constraint (2.5) ensures that the actions are taken with the probabilities specified by the policy $\pi$.

The return in (2.1) is maximized over policies that satisfy the quality constraints of the CMDP:

$$\sum_{s \in \mathcal{Q}_i} d_\pi(s) \leq q_i$$

for all $i$.

In the remainder of the paper, we use $\pi(s) = a$ to denote a deterministic policy that chooses $a$ with probability 1 and use $\pi(s, a)$ to denote a probability of taking an action $a$. Finally, $\pi(s)$ for a stochastic policy denotes the vector of action probabilities.

The constraints in the CMDP make it somewhat harder to solve than regular MDPs. In particular, the standard MDP solution methods, such as *value iteration* and *policy iteration* cannot be used. The main reason is that, as Remark 2.2 notes, the optimality of a policy depends on the initial distribution. Therefore, the optimal value function cannot be computed without a reference to the initial distribution. Constrained MDPs are instead solved using an extended linear program formulation of the MDP [2].

The CMDP with continuous probability modulations is even harder to solve than regular CMDPs because of the continuous action sets. In the remainder of the paper, we show how to solve the continuous-action CMDP when the reward function satisfies certain properties. In particular, if the rewards are affine the continuous-action CMDP can be reduced to one with a finite number of actions. More generally, when the rewards are concave there exists a tractable convex formulation and, surprisingly, there may exist a tractable formulation even when the rewards are non-concave.

The loan management problem can be formulated as a CMDP as follows. As mentioned above, we can formulate the evolution of each individual loan independently from other. Let the possible delinquency states be from a set $\mathcal{D}$. Assume, in addition, that the loan size is one of discrete levels from set $\mathcal{L}$; the value of loan may change as its state evolves and it is important in determining the cost of a default. The MDP states are then defined as:

$$\mathcal{S}_t = \{(t, s, l) \ : \ s \in \mathcal{D}, l \in \mathcal{L}\} \quad t = 1 \ldots T.$$

When no intervention is taken, the loan transitions between the states according to a base transition probability $b(s_t)$ for each $s_t \in \mathcal{S}_t$.

The transitions represent both the change in the delinquency state and the loan value. The interventions modify base transition probabilities to reduce the probability of the delinquency. The feasible actions we consider in our application are $\mathcal{A}(s_t) = \{p \in \Delta^{\mathcal{S}_{t+1}} \ : \ \|p - b(s_t)\|_\infty \leq \epsilon\}$—that is the difference from the base transition probability is bounded element-wise. Each intervention has a cost associated with it. The costs are convex in the scope of the transition probability modulation. In particular, we use an appropriately weighted version of $\|a - b(s_t)\|_1$ to represent the cost of action $a$ for each state $s_t$. The rewards correspond to negative costs and are, therefore, concave.

## 3 CMDPs with Affine Rewards

In this section, we show that the continuous action sets can be replaced by finite sets when 1) the rewards are affine functions of transition probabilities, and 2) the action sets $\mathcal{A}(s)$ are polytopes for every $s \in \mathcal{S}$. In particular, we show that there exists an optimal (randomized) policy that only takes actions that correspond to the extreme points of the polytope $\mathcal{A}(s)$.

**Assumption 1.** The reward $r(s, a)$ is an *affine* function of $a \in \mathcal{A}(s)$ for each $s \in \mathcal{S}$:

$$r(s, a) = e_s^\mathsf{T} a + f_s \, ,$$

for some $e_s$ and $f_s$.

Consider a CMDP $\mathcal{M}_1$ with continuous action sets as defined in Section 2. We can now construct a CMDP $\mathcal{M}_2$ with an identical state space to $\mathcal{M}_1$ and actions defined as:

$$\bar{\mathcal{A}}(s) = \text{ext}(\mathcal{A}(s)),$$

for each $s \in \mathcal{S}$ where $\text{ext}$ denotes the extreme points of the set. That is, the actions in $\mathcal{M}_2$ also define the actual transition probabilities as in $\mathcal{M}_1$; except the actions are restricted to the subset $\bar{\mathcal{A}}(s)$. The reward function $\mathcal{M}_2$ is identical to the reward function in $\mathcal{M}_1$.

**Theorem 3.1.** *Assume that $\mathcal{A}(s)$ is a convex polytope and that Assumption 1 holds. Then, the optimal returns in $\mathcal{M}_1$ and $\mathcal{M}_2$ are identical. In addition, for any optimal policy $\pi_2^\star$ in $\mathcal{M}_2$ there exists a deterministic policy $\pi_1^\star$ in $\mathcal{M}_1$ with the same return.*

To prove Theorem 3.1 we first need to establish the existence of an optimal deterministic policy for $\mathcal{M}_1$ when the reward function is concave (or affine).

**Lemma 3.2.** *Assume that the function $r(s, a)$ is concave in $a$ and $\mathcal{A}(s)$ is convex for each $s \in \mathcal{S}$. Then, there exists an optimal deterministic policy $\pi^\star$ in $\mathcal{M}_1$.*

*Proof.* Assume an optimal randomized policy $\pi_0 \in \Pi_R$; we show there exists a deterministic policy $\pi_1 \in \Pi_D$ such that $\rho(\pi_0) = \rho(\pi_1)$. The deterministic policy $\pi_1$ is constructed as:

$$\pi_1(s) = \sum_{a \in \mathcal{A}(s)} \pi_0(s, a) \cdot a,$$

for each $s \in \mathcal{S}$. Note that $a$ is vector in this equation; that is the action $\pi_1$ is a convex combination of elements of $\mathcal{A}(s)$.

The action $\pi_1(s)$ is in $\mathcal{A}(s)$ from because this is a convex set and $\pi_1(s)$ is a convex combination of the elements of the set. Using (2.3) and (2.4) the state visitation probabilities of $\pi_1$ and $\pi_2$ are the same: $d_{\pi_0} = d_{\pi_1}$. Using this equality and the concavity of $r$, we have that:

$$r_{\pi_1}(s) = r(s, \pi_1(s)) = r\Big(s, \sum_{a \in \mathcal{A}(s)} \pi_0(s, a) \cdot a\Big)$$
$$\geq \sum_{a \in \mathcal{A}(s)} \pi_0(s, a) r(s, a) = r_{\pi_0}(s).$$

It readily follows that the transition probabilities under $\pi_0$ and $\pi_1$ are identical and therefore $\rho(\pi_1) \geq \rho(\pi_0)$. The lemma then follows from the optimality of $\pi_0$ and from the monotonicity of the Bellman operator. The monotonicity of the Bellman operator implies that uniformly increasing the rewards also increases the return. □

*Proof of Theorem 3.1.* Let $\pi_i^\star$ and $\rho_i^\star$ be the optimal policy and return in $\mathcal{M}_i$ respectively. We show the equality $\rho_1^\star = \rho_2^\star$ in two steps; first, we show that $\rho_2^\star \geq \rho_1^\star$. Assume, from Lemma 3.2, that $\pi_1^\star$ is deterministic. Then, create a *randomized* policy $\pi_2$ in $\mathcal{M}_2$ such that for each $s \in \mathcal{S}$ it satisfies:

$$\sum_{\bar{a} \in \hat{\mathcal{A}}(s)} \pi_2(s, \bar{a}) \cdot \bar{a} = \pi_1^\star(s) \qquad (3.1)$$

There always exists a unique $\pi_2$ that satisfies the above condition since $\hat{\mathcal{A}}(s) = \text{ext}(\mathcal{A}(s))$ and $\mathcal{A}(s)$ is convex—each point in a polytope is a unique convex combination of its extreme points (e.g. Krein–Milman Theorem). The condition (3.1) guarantees that the transitions probabilities for $\pi_1^\star$ and $\pi_2$ are the same. It remains to show that the rewards for $\pi_1^\star$ and $\pi_2$ equal:

$$\begin{aligned} r_{\pi_2}(s) &= \sum_{\bar{a} \in \hat{\mathcal{A}}(s)} \pi_2(s, \bar{a}) \cdot r(s, \bar{a}) \\ &= e_s^\mathsf{T} \Big( \sum_{\bar{a} \in \hat{\mathcal{A}}(s)} \pi_2(s, \bar{a}) \cdot \bar{a} \Big) + f_s = r_{\pi_1^\star}(s) \end{aligned} \qquad (3.2)$$

by (3.1) of $\pi_2$ and Assumption 1. The monotonicity of the Bellman operator then implies that $\rho_2^\star \geq \rho_2 \geq \rho_1^\star$.

Next, we show that $\rho_1^\star \geq \rho_2^\star$. Let $\pi_2^\star$ be an optimal *randomized* policy in $\mathcal{M}_2$. Define a deterministic policy $\pi_1$ as follows:

$$\pi_1(s) = \sum_{\bar{a} \in \hat{\mathcal{A}}(s)} \pi_2^\star(s, \bar{a}) \cdot \bar{a}. \qquad (3.3)$$

Note that (3.3) represents a convex combination of individual action vectors. It can be readily shown from (3.3) that the transition probabilities for policies $\pi_2^\star$ and $\pi_1$ are the same. Next, we show that $r_{\pi_1}(s) \geq r_{\pi_2^\star}(s)$:

$$\begin{aligned} r_{\pi_2^\star}(s) &= \sum_{\bar{a} \in \hat{\mathcal{A}}(s)} \pi_2^\star(s, \bar{a}) \cdot r(s, \bar{a}) \\ &\leq r\Big(s, \sum_{\bar{a} \in \hat{\mathcal{A}}(s)} \pi_2^\star(s, \bar{a}) \cdot \bar{a}\Big) \\ &= r(a, \pi_1(s)) = r_{\pi_1}(s), \end{aligned}$$

using the concavity of the reward function. The monotonicity of the Bellman operator implies that $\rho_1^\star \geq \rho_2^\star$. This shows the required equality and the necessary policies can be constructed as defined in (3.2) and in (3.3). □

There are two main limitations of the reduction in Theorem 3.1. First, the reward function must be linear. This limitation can be easily relaxed by extending the results to rewards that are piece-wise linear and concave by considering the extreme points of the hypograph of this function. Second, even though the number of actions in this formulation is finite, it still may be very large; in the worst case, the number of the finite actions may be exponential in the number of states even when $\mathcal{A}$ are specified by a polynomial number of linear constraints. In the following sections, we resolve this limitation by directly formulating the continuous-action CMDP as a convex optimization problem.

## 4 CMDPs with Concave Rewards

In this section, we describe a direct formulation of the CMDP as a convex mathematical optimization problem. This formulation significantly relaxes the necessary assumptions on the MDP structure compared to Theorem 3.1 and also leads to a tractable algorithm.

We start by extending the reward function $r : \mathcal{S}_t \times \Delta^{|\mathcal{S}_{t+1}|} \to \mathbb{R}$ to $\bar{r} : \mathcal{S}_t \times \mathbb{R}_+^{|\mathcal{S}_{t+1}|} \to \mathbb{R}$ which also assigns rewards for actions that are not valid distributions. The extended function $\bar{r}(s, a)$ is defined as:

$$\bar{r}(s, a) = \mathbf{1}^\mathsf{T} a \cdot r\left(s, \frac{a}{\mathbf{1}^\mathsf{T} a}\right),$$

where $\bar{r}(s, \mathbf{0}) = 0$. Note that this function is positively homogeneous; that is $\bar{r}(s, q \cdot a) = q \cdot r(s, a)$ for $q \geq 0$. This transformation also preserves the convexity or concavity of the reward function as the following lemma states.

**Lemma 4.1.** *For each $s_t \in \mathcal{S}_t$, the function $\bar{f}(a) = \mathbf{1}^\mathsf{T} a \cdot f(a/\mathbf{1}^\mathsf{T} a)$ is concave (convex) on $\mathbb{R}^{|\mathcal{S}_{t+1}|}$ if and only if $f(a)$ is concave (convex) on $\Delta^{|\mathcal{S}_{t+1}|}$.*

*Proof.* This is a standard result which can be readily shown directly from the definition of concavity (convexity) for $q \cdot f(x/q)$ for $q \geq 0$. Assume any non-negative $\alpha + \beta = 1$, then:

$$(\alpha q_1 + \beta q_2) \cdot f\left(\frac{\alpha x_1 + \beta x_2}{\alpha q_1 + \beta q_2}\right) =$$
$$= (\alpha q_1 + \beta q_2) \cdot f\left(\frac{\alpha x_1 q_1}{\alpha q_1 + \beta q_2} \frac{x_1}{q_1} + \frac{\beta x_2 q_2}{\alpha q_1 + \beta q_2} \frac{x_2}{q_2}\right) =$$
$$= \alpha q_1 \cdot f\left(\frac{x_1}{q_1}\right) + \beta q_2 \cdot f\left(\frac{x_2}{q_2}\right).$$

The lemma then follows from the restriction of $q = \mathbf{1}^\mathsf{T} x$. □

Below, we show several examples of the extended function.

**Example 4.2.** *Assume that the reward is linear: $r(s, a) = e_s^\mathsf{T} a + f_s$. Then, the extended reward function is:*

$$\bar{r}(s, a) = e_s^\mathsf{T} a + \mathbf{1}^\mathsf{T} a \cdot f_s .$$

**Example 4.3.** *Assume that the reward is defined by a norm: $r(s, a) = -\|a - \bar{a}_s\|$. Then, the extended reward function is:*

$$\bar{r}(s, a) = -\|a - \mathbf{1}^\mathsf{T} a \cdot \bar{a}_s\| .$$

**Example 4.4.** *Assume that the reward is defined by a squared $L_2$ norm: $r(s, a) = -\|a - \bar{a}_s\|_2^2$. Then, the extended reward function is:*

$$\bar{r}(s, a) = -\frac{1}{\mathbf{1}^\mathsf{T} a} \cdot \|a - \mathbf{1}^\mathsf{T} a \cdot \bar{a}_s\|_2^2 .$$

We are now ready to formulate the convex optimization problem. Constrained MDPs are typically solved using a linear program formulation based on the state-action visitation probabilities $u$ as the optimization variables [2]. Such formulation would clearly lead to a semi-infinite optimization problem because of the continuous action space and the need to have a decision variable for each state and action pair. To get a tractable formulation, we instead use decision variables $u(s_t, s_{t+1})$, which represent the *joint* probability of visiting $s_t$ *and* transiting to $s_{t+1}$. State visitation probabilities $d(s_t)$ can be derived from these variables by marginalizing over $s_{t+1}$ similar to (2.2).

The main challenge with the formulation based on the decision variables $u(s_t, s_{t+1})$ is to ensure that the corresponding transition probabilities represent feasible actions in $\mathcal{A}(s)$. We use the notation $u(s_t, \cdot)$ represents the vector of values indexed by the second argument. Then, the vector of transition probabilities from state $s_t$ is $u(s_t, \cdot)/d(s_t)$ which must be feasible in $\mathcal{A}(s_t)$. The constraints $u(s_t, \cdot)/d(s_t) \in \mathcal{A}(s_t)$ are non-linear and non-convex in the state visitation probabilities $d(s_t)$. Therefore, a direct formulation would be non-convex and difficult to solve.

To derive a convex formulation, let $\mathcal{A}(s_t)$ be a convex set defined by *convex* constraints for $s_t \in \mathcal{S}_t$:

$$\mathcal{A}(s_t) = \{a \in \Delta^{|\mathcal{S}_{t+1}|} \,:\, f_{s_t}^j(a) \leq 0, j \in \mathcal{J}\} ,$$

for some $f_{s_t}^j$. The feasibility constraints on the transition probabilities that have to be satisfied by the solution $u$ then become:

$$f_s^j\left(\frac{u(s, \cdot)}{d(s)}\right) \leq 0 . \qquad (4.1)$$

This function is non-convex in $d(s)$ and, therefore, cannot be used to formulate a convex optimization problem. To get an identical but convex constraint, first define an extended constraint function:

$$\bar{f}_s^j(a) = \mathbf{1}^\mathsf{T} a \cdot f_s^j\left(\frac{a}{\mathbf{1}^\mathsf{T} a}\right),$$

where by definition $\bar{f}_s^j(\mathbf{0}) = 0$. Note that $d(s) = \mathbf{1}^\mathsf{T} u(s, \cdot)$. The constraint (4.1) can be multiplied by $d(s)$ to get the constraint:

$$d(s) \cdot f_s^j\left(\frac{u(s, \cdot)}{d(s)}\right) = \bar{f}_s^j(u(s, \cdot)) \leq 0 . \qquad (4.2)$$

The function $\bar{f}_s^j$ is convex from Lemma 4.1 and the constraint (4.2) is equivalent to (4.1) since $d(s) \geq 0$ and $u(s, \cdot) = \mathbf{0}$ whenever $d(s) = 0$.

We are now ready to formulate the optimization problem

that can be used to compute the optimal policy in CMDPs:

$$\begin{aligned}
\max_{u \geq 0, d \geq 0} \quad & \sum_{s \in \mathcal{S}} \bar{r}(s, u(s, \cdot)) \\
\text{s.t.} \quad & d(s_1) = \alpha(s_1) \quad \forall s_1 \in \mathcal{S}_1 \\
& d(s_t) = \sum_{s_{t+1}} u(s_t, s_{t+1}) \\
& d(s_t) = \sum_{s_{t-1}} u(s_{t-1}, s_t) \\
& \sum_{s \in \mathcal{Q}_i} d(s) \leq q_i \quad i \in \mathcal{I} \\
& \bar{f}_s^j(u(s, \cdot)) \leq 0 \quad j \in \mathcal{J}
\end{aligned} \quad (4.3)$$

Each $s_t$ is implicitly considered to be in $\mathcal{S}_t$ and each $s$ is implicitly considered to be in $\mathcal{S}$. Note that:

$$\sum_{s \in \mathcal{S}} \bar{r}(s, u(s, \cdot)) = \sum_{s \in \mathcal{S}} d(s) \cdot r\left(s, \frac{u(s, \cdot)}{d(s)}\right).$$

The formulation in (4.3) reduces to a linear program when the sets of feasible actions are polytopes as the following example shows.

The intuitive meaning of the constraints in (4.3) the same as in Eqs. (2.2) to (2.5). The main difference from the standard LP formulation is the objective function, which is expressed in terms of the extended reward function, and the last constraint, which is expressed in terms of the extended action constraint functions. The optimal policy $\pi^\star$ can be extracted from the optimal solution $u^\star, d^\star$ as according to Theorem 4.6.

**Example 4.5.** *Assume that the set of feasible actions is a polytope for each $s_t \in \mathcal{S}_t$:*

$$\mathcal{A}(s_t) = \{a \in \Delta^{|\mathcal{S}_{t+1}|} \ : \ H_s a \leq h_s\}.$$

*Then, the constraints $\bar{f}_s^j(a) \leq 0$ for all $j \in \mathcal{J}$ become:*

$$\begin{aligned}
\mathbf{1}^\mathsf{T} a \cdot H_s \frac{a}{\mathbf{1}^\mathsf{T} a} &\leq \mathbf{1}^\mathsf{T} a \cdot h_s \\
H_s a &\leq \mathbf{1}^\mathsf{T} a \cdot h_s \, ,
\end{aligned}$$

*which is a set of linear constraints.*

The following theorem states the correctness of the formulation (4.3).

**Theorem 4.6.** *Assume that, for each $s \in \mathcal{S}$, $r(s, a)$ is concave in $a$ and the set $\mathcal{A}(s)$ is convex. Let $u^\star, d^\star$ be the optimal solution of* (4.3) *and define a* deterministic *policy $\pi$:*

$$\pi(s) = u^\star(s, \cdot)/d^\star(s).$$

*That is, $\pi(s)$ maps a state to a vector of state transition probabilities. Then, $\pi$ is an optimal policy and the objective value of* (4.3) *equals to $\rho(\pi)$. In addition,* (4.3) *is a convex optimization problem.*

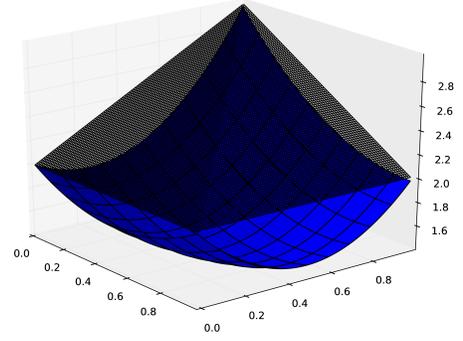

Figure 1: A convex function and its concave envelope over a unit square.

*Proof.* We first show that the optimal policy $\pi^\star$ is feasible in (4.3) and the corresponding objective value equals the return of the optimal policy $\pi^\star$. Given an optimal *deterministic* policy $\pi^\star$ (from Lemma 3.2), construct the solution $u, d$ in (4.3) as $u(s_t, \cdot) = d(s_t) \cdot \pi^\star(s_t, \cdot)$. It is well known (e.g. [9]) that there is a unique such solution to all constraints without $\bar{f}_s^j(u(s, \cdot)) \leq 0$. As described above, this constraint is valid from (4.1) and (4.2) because $d \geq \mathbf{0}$. Therefore, $\sum_{s \in \mathcal{S}} \bar{r}(s, u^\star(s, \cdot)) \geq \rho(\pi^\star)$. The reverse inequality $\sum_{s \in \mathcal{S}} \bar{r}(s, u^\star(s, \cdot)) \leq \rho(\pi^\star)$ can be shown similarly by constructing a feasible policy from any solution $u, d$ using the construction from the statement of the theorem. The convexity of the optimization problem follows readily from Lemma 4.1. □

The computational complexity of solving (4.3) depends on the form of $\bar{r}$; the problem is tractable for most common concave functions. In particular, (4.3) is tractable for concave piecewise linear functions and concave quadratic functions. Note that this formulation generalizes the setting in Section 3 and has a smaller computational complexity.

## 5 CMDPs with Non-concave Rewards

In this section, we describe how to tractably solve CMDPs with non-concave reward functions. The approach relies on the fact that the optimal return of any constrained MDP is unaffected if the rewards are replaced by their *concave envelope* thereby obtaining a concave maximization problem.

The *concave envelope* $g(x)$ of a function $f(x)$ is defined as [3]:

$$g(x) = \sup\{t \ : \ (x, t) \in \operatorname{conv} \operatorname{hypo} f\},$$

where $\operatorname{conv}$ is the convex hull and $\operatorname{hypo}$ is the hypograph of $f$. A hypograph of $f$ is defined as: $\operatorname{hypo} f = \{(x, t) \ : \ t \leq f(x)\}$. The supremum above is achieved whenever $f$ is bounded and $\mathcal{A}(s)$ are compact, which are the assumptions that we make. A concave envelope is important because it is the smallest concave function that is greater than $f$.

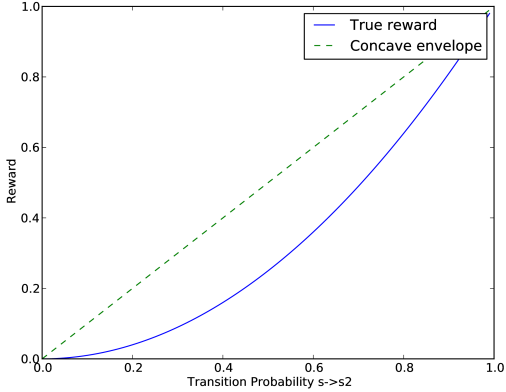

Figure 2: Example of concave envelope of the reward achievable by a randomized policy.

**Example 5.1.** *Consider a function $f(x,y) = x^2 + 2 \cdot y^2 - x \cdot y + 2 - x - y$ defined on the interval $[0,1] \times [0,1]$. The concave envelope of this convex function is the piecewise linear concave function $g(x,y) = \min\{y+2, -x+3\}$. Fig. 1 shows the convex function $f$ and its concave envelope $g$.*

Assume a CMDP $\mathcal{M}$ with a reward function $r$ and construct a CMDP $\mathcal{M}_e$ with a reward $r^e$ that is the *concave envelope* of $r$ for each $s \in \mathcal{S}$:

$$r^e(s,a) = \sup\{t \;:\; (x,t) \in \text{conv hypo}\, r(s,x)\},$$

where hypo is over $\mathcal{A}(s)$. Let $\rho(\pi)$ and $\rho_e(\pi)$ be the returns of $\pi$ in $\mathcal{M}$ and $\mathcal{M}_e$ respectively.

The motivation for considering the concave envelope of the rewards is that the transition probabilities with this reward can be actually achieved by appropriately randomizing the policy. The following example shows this property.

**Example 5.2.** *Consider a state $s$ with transitions to two other states $s_1$ and $s_2$ with continuous modulation of probabilities in the set $\mathcal{A}(s) = \Delta^2$. For any action $a$, let $a_1$ and $a_2$ represent the transition probabilities to states $s_1$ and $s_2$ respectively. Consider a* convex *reward function $r(s,a) = a_2^2$ and its concave envelope $r_e(s,a) = a_2$ depicted in Fig. 2. To show that the optimal policy will be always randomized between the extreme points, assume for example that the optimal policy is to take the transition probability $(0.6, 0.4)$. Directly taking an action $(0.6, 0.4)$ accrues a reward $0.4^2 = 0.16$. However, taking action $(0,1)$ with probability $0.4$ and action $(1,0)$ with probability $0.6$ accrues a higher reward of $0.4$. In general, the maximal reward for each transition probability can be achieved by the maximal convex combination of other feasible actions which exactly yields the concave envelope.*

The CMDP $\mathcal{M}$ cannot be solved using (4.3) because of the non-concave rewards. On the other hand, because the rewards in $\mathcal{M}_e$ are concave, it can be easily formulated as (4.3). Note, however, that the optimal solution of $\mathcal{M}_e$ is not necessarily optimal in $\mathcal{M}$. The following theorem states that the optimal solution for $\mathcal{M}$ can be easily constructed from the optimal solution to $\mathcal{M}_e$ by appropriately randomizing between the extreme points of the concave envelope.

**Theorem 5.3.** *Let $\pi_e^\star$ be an optimal policy in CMDP $\mathcal{M}_e$. Then, one can construct an optimal policy $\pi^\star$ in $\mathcal{M}$ such that 1) $\rho_e(\pi_e^\star) = \rho(\pi^\star)$ and 2) the transition probabilities $\pi^\star$ and $\pi_e^\star$ are identical.*

*Proof.* First, we can assume $\pi_e^\star$ to be deterministic without loss of generality from Lemma 3.2. Clearly, we have from the optimality of $\pi_e^\star$ and from $r_e(s,a) \geq r(s,a)$ that:

$$\rho_e(\pi_e^\star) \geq \rho_e(\pi^\star) \geq \rho(\pi^\star).$$

To show the equality, it only remains to show that $\rho_e(\pi_e^\star) \leq \rho(\pi^\star)$. For any $s \in \mathcal{S}$, because the value $r^e(s,\cdot)$ is a maximum in a *closed* convex hull, it is on its boundary. Therefore, for any $a$ there exist $a_i \in \mathcal{A}(s)$ such that $r^e(s, a_i) = r(s, a_i)$ (i.e. the extreme points of the hypograph) and $\lambda_i \in [0,1]$ such that:

$$r^e(s,a) = \sum_{i=1}^{m} \lambda_i \cdot r(s, a_i),$$

such that $\lambda \geq \mathbf{0}$, $\sum_i \lambda_i = 1$, and $a = \sum_i \lambda_i \cdot a_i$. Then, construct a policy $\pi$ as follows:

$$\pi(s, a_i) = \lambda_i.$$

It can be shown readily that the transition probabilities of $\pi$ and $\pi_e^\star$ are the same, since $a = \sum_i \lambda_i \cdot a_i$ when assuming $a = \pi_e^\star(s)$. Then:

$$r_\pi(s) = \sum_i \lambda_i \cdot r(s, a_i) = r^e(s,a) = r^e_{\pi_e^\star}(s).$$

Therefore, the rewards and transitions of $\pi$ and $\pi^\star$ are the same, which also implies $\rho_e(\pi_e^\star) \leq \rho(\pi^\star)$. □

A CMDP with non-concave rewards, therefore, can be solved as follows. First, construct a concave envelope of the rewards. Then, use (4.3) to solve the new CMDP and get a policy $\pi_e^\star$. Finally, construct the optimal $\pi^\star$ according to the construction in the proof of Theorem 5.3. That is, any action $a$ is replaced by randomizing among actions $a_i$ by probabilities $\lambda_i$. The points $a_i$ depend on the construction of the concave envelope. The values $\lambda_i$ can be readily computed by linear programming in general settings.

The tractability of the concave envelope approach depends on several factors. First, constructing a concave envelope is difficult in general. Second, the computed concave envelope may not have a formulation that is easily optimized. A particular case of interest is when the rewards are *convex*. Then, the concave envelope is piecewise linear and can be expressed in terms of the extreme points of $\mathcal{A}(s)$ as a linear program—it is a maximization over the convex combination of the extreme points. When the reward function is submodular on the lattice of extreme points, the envelope can be further simplified [10].

# 6 Application to Loan Delinquency Management

In this section, we describe the empirical results from an application of the new CMDP solution methods for both a real and a synthetic loan delinquency management problem.

We applied the proposed methods to managing the delinquencies of a loan portfolio of an actual service provider. While we are not authorized to disclose detailed results of this application, we can report the impact of our solution method. There are 8 possible states of loan delinquency; the transition probabilities can be modulated in 4 of them. The probabilities are influenced by investing resources, such as principal reduction, in the appropriate loans. The portfolio performance targets need to be achieved within a horizon of 6 months. The ranges of possible modulations and their costs were derived from corresponding transition probabilities in prior months.

The real-world empirical study was conducted to establish the necessity of a global optimization method for solving this problem. We initially evaluated a simple greedy algorithm which iteratively finds an optimal modulation of probabilities in a month $t$ assuming that the base transitions in future months will not be modified. This greedy method returned solutions characterized by high fluctuations in monthly investments in loan servicing operations. Because the method assumes no modulations after the month $t$, the modulations in month $t$ had to be overly aggressive. In the next month $t + 1$, the portfolio would be in a sufficiently good state to merit no further modulations. These month-to-month fluctuation are resource-intensive and undesirable. The optimal method proposed here smoothens out these fluctuations and can result in a significant overall reduction of resources needed to meet portfolio targets over the whole planning horizon. Experiments on six actual loan portfolios for a time horizon of six months have revealed that using the optimal method proposed in this paper has allowed for an average 13.97% reduction in the expected costs of portfolio servicing operations in comparison with the benchmark strategies used by loan managers.

Next, we proceed with an evaluation of the solution quality and scalability of the proposed algorithms on a set of synthetic loan delinquency management problems. We consider a variable number of loan delinquency states and a fixed horizon of 6 periods. The states are ordered; the increasing order represent the increasing delinquency state of a loan, such as the number of weeks behind payments. The first state represents the loan to be current and the last state represents the default. The probability of increasing delinquency in the given period increases logarithmically with the current state of delinquency. In other words, accounts that are delinquent now are more likely to become even more delinquent in the future. The probability of the delinquency decreasing to any less delinquent state is uniform. The feasible actions are allowed to modulate any single transition probability by at most $\epsilon = 0.4$. The rewards are linear in the deviation from the base probability in each element: $-\|a - b\|_1$. The quality constraints on the probability of the default (last state) is $q = 0.04$.

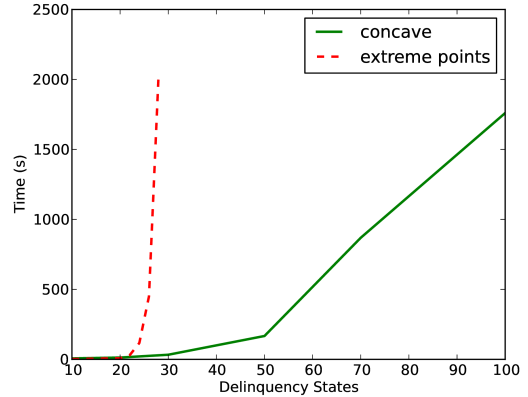

Figure 3: Time to solve a CMDP for as a function of number of states. The method "extreme points" is described in Section 3 and "concave" is described in Section 4.

Fig. 3 compares the time to solve the CMDP using the extreme points formulation described in Section 3 versus the tractable concave method described in Section 4. The timings were obtained using CPLEX 12.5 running on an Intel Core i5 1.5 GHz processor. As expected, the tractable method scales much better with the number of states. While the concave method can easily solve problems with 100s of states, the extreme-point method becomes intractable with more than 30 states. In our benchmark problem, the number of extreme points grows exponentially with the number of states. The solution quality with the two methods is identical since they are both optimal. Fig. 4 shows the sensitivity of the return to the quality constraint—the limit on the probability of a loan to end in default.

Because the rewards may often be non-concave, we also evaluate the approach assuming convex quadratic rewards $\|a - b\|_2^2$; this is relevant in particular when the economies of scale become important. We compare the algorithm from Section 5 with a simple naive approach which uses

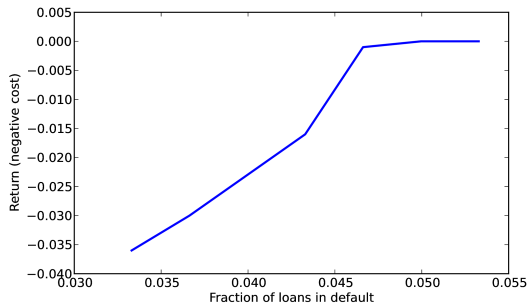

Figure 4: Return of the optimal solution as a function of the limit on the fraction of loans in default.

a linear approximation of the reward function. For the quadratic function and 30 delinquency states, the optimal method based on the concave envelope achieves a return of 0.14, while the approximation achieves return of 0. The difference in the return between these two methods can be arbitrarily large depending on the problem formulation.

## 7 Conclusion

We proposed three solution methods for solving constrained MDPs with continuous modulation of the probabilities. The MDP formulation was motivated by a practical need to optimally manage the delinquencies of a loan portfolio. We are not aware of any previous methods in the literature that can be used to solve this class of problems. The first method reduces the continuous action sets to finite when the rewards are affine and feasible sets polyhedral. The second formulation is a tractable optimization problem which applies to arbitrary concave reward functions. Finally, the third formulation extends the second one to non-concave rewards.

Our experimental results show that the method based on the convex optimization problem scales well and can solve problems with a large number of states in a few seconds. The method based on extreme point enumeration does not scale well, but performs better for very small problems and can be used in theoretical analysis of the result. Finally, when using the concave envelope of the rewards can significantly improve the solution quality when compared to naive approaches. While this method has not been deployed yet, the initial test results indicate that it can lead to significant improvements compared with the current greedy approach.

There are several important way in which our results can be extended. First, we considered a risk-neutral loan service provider whose utility can be expressed in terms of expectations. However, it may be desirable to extend the approach to risk-averse setting in which the service provider would be willing to trade off a higher servicing cost for a lower probability of violating the quality constraints. Other extensions involve improving the scalability of the concave envelopes for various classes of convex functions and extensions to problems with many states.